\documentclass[dvipsnames]{article} %
\usepackage{iclr2021_conference,times}

\usepackage{amsmath,amsfonts,bm}

\def\eqref#1{equation~\ref{#1}}

\def\1{\bm{1}}

\DeclareMathAlphabet{\mathsfit}{\encodingdefault}{\sfdefault}{m}{sl}
\SetMathAlphabet{\mathsfit}{bold}{\encodingdefault}{\sfdefault}{bx}{n}

\DeclareMathOperator*{\argmax}{arg\,max}

\usepackage{amsmath}

\usepackage{hyperref}
\usepackage{url}

\usepackage{nccmath} %

\usepackage{booktabs}       %

\usepackage{pgfplots, amsthm, graphicx, subcaption, dsfont, tikz, tikz-qtree, tabularx, amssymb,enumerate, verbatim, enumitem,makecell,authblk}%
\usepackage[colorinlistoftodos, textwidth=20mm]{todonotes}%

\usepackage{float} 
\usepackage[bottom]{footmisc} %
\usepackage{footnote}
\usepackage[noabbrev, capitalise]{cleveref} %

\usepackage{xcolor} %

\pgfplotsset{compat=1.11} %
\usepgfplotslibrary{fillbetween}
\usepgfplotslibrary{groupplots}%

\usepackage[switch]{lineno} %

\title{Measuring Adversarial Robustness using a Voronoi-Epsilon Adversary} %

\author[1,2]{Hyeongji Kim} %
\author[1]{Pekka Parviainen}
\author[1,2]{Ketil Malde}
\affil[1]{Department of Informatics, University of Bergen, Norway}
\affil[2]{Institute of Marine Research, Bergen, Norway}
\affil[ ]{\texttt{kim.hyeongji@hi.no}}

\iclrfinalcopy %
\begin{document}

\maketitle

\begin{abstract}
Previous studies on robustness have argued that there is a tradeoff between accuracy and adversarial accuracy. The tradeoff can be inevitable even when we neglect generalization. We argue that the tradeoff is inherent to the commonly used definition %
of adversarial accuracy, which uses 
an adversary that can construct adversarial points constrained by \(\epsilon\)-balls around data points. 
As \(\epsilon\) gets large, 
the adversary may %
use real data points from other classes as adversarial examples. 
We propose a Voronoi-epsilon adversary %
which is constrained both by Voronoi cells and by \(\epsilon\)-balls. %
This adversary balances between two notions of perturbation. %
As a result, adversarial accuracy based on this adversary avoids a tradeoff between accuracy and adversarial accuracy on training data even when \(\epsilon\) is large. %
Finally, we show that a nearest neighbor classifier is the maximally robust classifier against the proposed adversary on the training data.
\end{abstract}

\newtheorem*{my_pro_setting}{Problem setting}

\newtheorem*{my_notation}{Notation}

\newtheorem{mydef}{Definition}

\section{Introduction}

By applying a carefully crafted, but imperceptible perturbation to input images, so-called adversarial examples can be constructed that cause classifiers to misclassify the perturbed inputs \citep{szegedy2013intriguing}. Defense methods like adversarial training \citep{madry2017towards} and certified defenses \citep{wong2018provable} %
against adversarial examples have often resulted in decreased accuracies on clean samples \citep{tsipras2018robustness}. Previous studies %
have argued that the tradeoff between accuracy and adversarial accuracy may be inevitable in classifiers \citep{tsipras2018robustness,dohmatob_limitations_2018,zhang2019theoretically}.

\subsection{Problem Settings}

\begin{my_pro_setting}\label{setting1}
Let \(\mathcal{X} \subset \mathbb{R}^{\dim}\) be a nonempty input space and \(\mathcal{Y}\) be a set of possible classes. %
Data points \(x\in\mathcal{X}\) and corresponding classes \(c_x\in\mathcal{Y}\) are sampled from a joint distribution \(\mathcal{D}\). %
The distribution \(\mathcal{D}\) should satisfy the condition that \(c_x\) is unique for all \(x\). The set of the data points is denoted as \(X\). \(X\) is a nonempty finite set. 
A classifier \(f\) assigns a class label from \(\mathcal{Y}\) for each point \(x \in \mathcal{X}\). 
\(L(x,y)\) is a classification loss of the classifier \(f\) provided an input \(x\in \mathcal{X}\) and a label \(y \in \mathcal{Y}\). %
\end{my_pro_setting}
More notations are summarized in \ref{sec:notaion_list}. Abbreviations are summarized in \ref{sec:abbreviation_list}. We %
focus on situations that we neglect generalization to simplify the analysis.%

\subsection{Adversarial Accuracy (AA)}
Adversarial accuracy is a commonly used measure of adversarial robustness of classifiers \citep{madry2017towards, tsipras2018robustness}. It is defined by an adversary region \(R(x)\subset \mathcal{X}\), which is an allowed region of the perturbations for a data point \(x\).

\begin{mydef}[\textbf{Adversarial accuracy}]\label{def:adv_acc}
Given an adversary that is constrained to an adversary region \(R({x})\), %
adversarial accuracy $a $ %
is defined as follows.
\begin{equation*}
    a=\mathbb{E}_{(x,c_x)\sim \mathcal{D}} \left[ {  \mathds{1} \left( f (x^{*})=c_x \right) } \right] %
    \text{ where } x^{*}=\argmax\limits_{x'\in R({x})}{L(x',c_x)}%
\end{equation*}
\end{mydef}

The choice of \(R({x})\) will determine the adversarial accuracy that we are measuring. Commonly considered adversary region is \(\mathbb{B}(x,\epsilon)\), which is a \(\epsilon\)-ball around a data point \(x\) based on a distance metric \(d\) \citep{biggio2013evasion, madry2017towards, tsipras2018robustness, zhang2019theoretically}.

\begin{mydef}[\textbf{Standard adversarial accuracy}]\label{def:std_adv_acc}
When the adversary region is \(\mathbb{B}(x,\epsilon)\), we refer to the adversarial accuracy $a$  %
as standard adversarial accuracy (SAA) $a_{std} (\epsilon)$. For SAA, we denote \(R({x})\) as \(R_{std}({\epsilon;x})\). %
\begin{equation*}
    a_{std} (\epsilon)=\mathbb{E}_{(x,c_x)\sim \mathcal{D}} \left[ {  \mathds{1} \left( f (x^{*})=c_x \right) } \right] %
    \text{ where } x^{*}=\argmax\limits_{x'\in R_{std}({\epsilon;x})}{L(x',c_x)}
\end{equation*}
\end{mydef}
This %
adversary region \(\mathbb{B}(x,\epsilon)\) is based on an implicit assumption that there might be an adequate single epsilon \(\epsilon\) that perturbed samples do not change their classes. %
However, this assumption has some limitations. We explain that in the next section. 

\subsection{The Tradeoff Between Accuracy and Standard Adversarial Accuracy}

The usage of \(\epsilon\)-ball-based adversary can cause the tradeoff between accuracy and adversarial accuracy. %
When the two clean samples $x_1$ and $x_2$ with \(d(x_1,x_2)\le\epsilon\) have different classes, the increase of standard adversarial accuracy requires misclassification. We illustrate this with a toy example.%

\subsubsection{Toy Example}\label{sec:toy_ex}
\begin{figure}
    \centering
    \begin{subfigure}[b]{0.25\linewidth} %
        \includegraphics[width=1\linewidth]{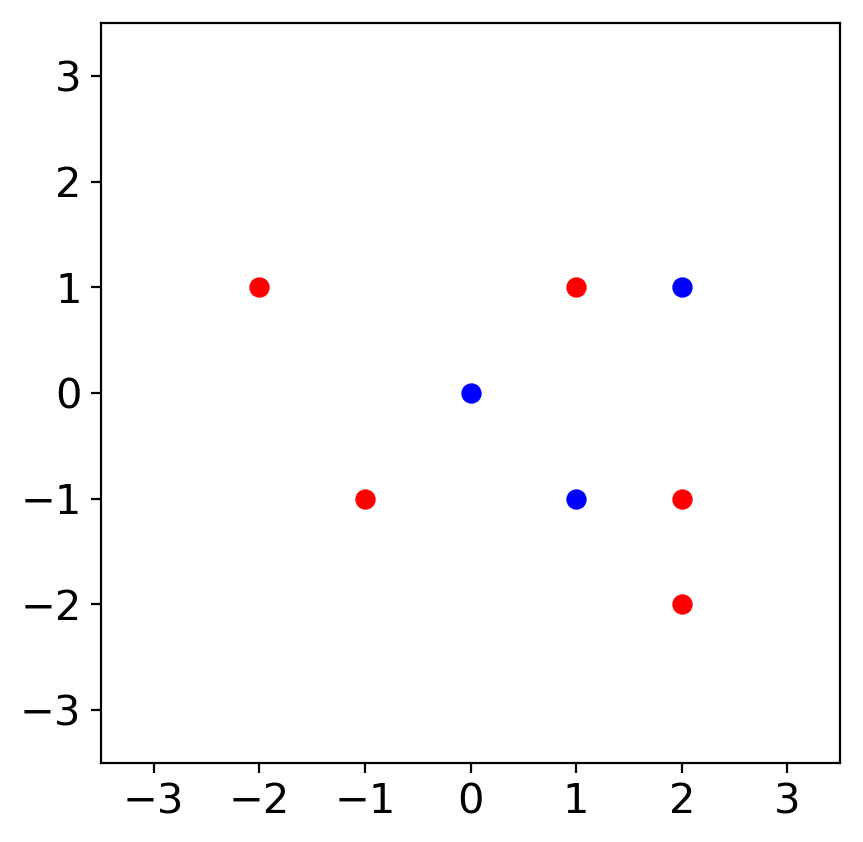}
        \caption{{}}\label{fig:toy_ex_SAA_a}
    \end{subfigure}
    \begin{subfigure}[b]{0.25\linewidth}%
        \includegraphics[width=1\linewidth]{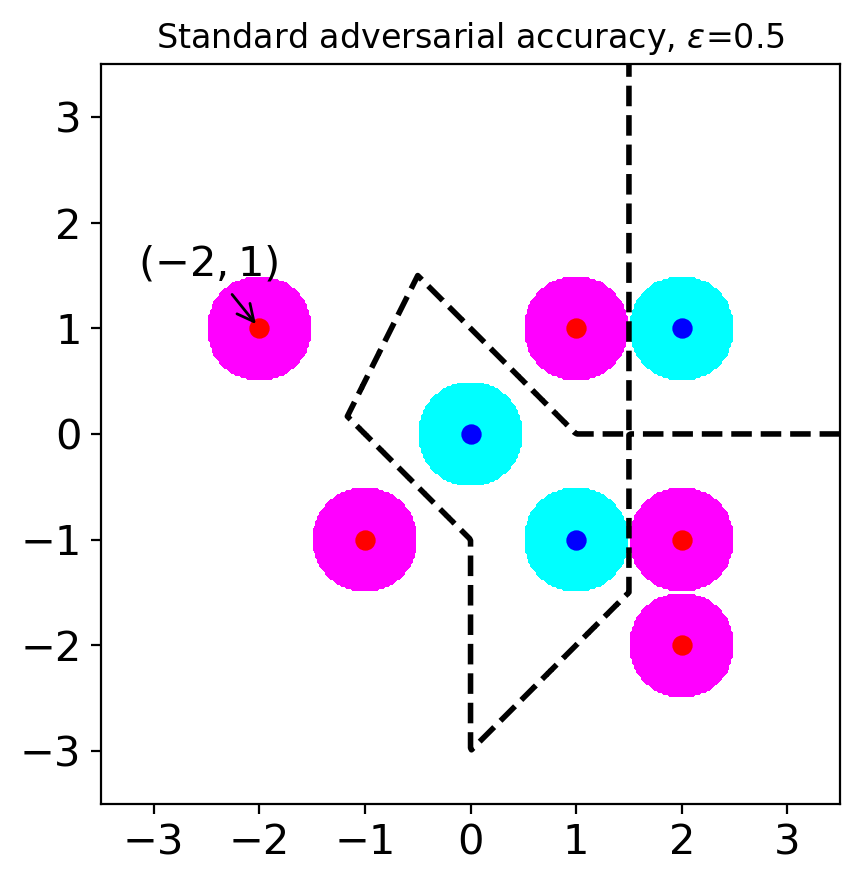} 
        \caption{{}}\label{fig:toy_ex_SAA_b}
    \end{subfigure}
    \begin{subfigure}[b]{0.25\linewidth}%
        \includegraphics[width=1\linewidth]{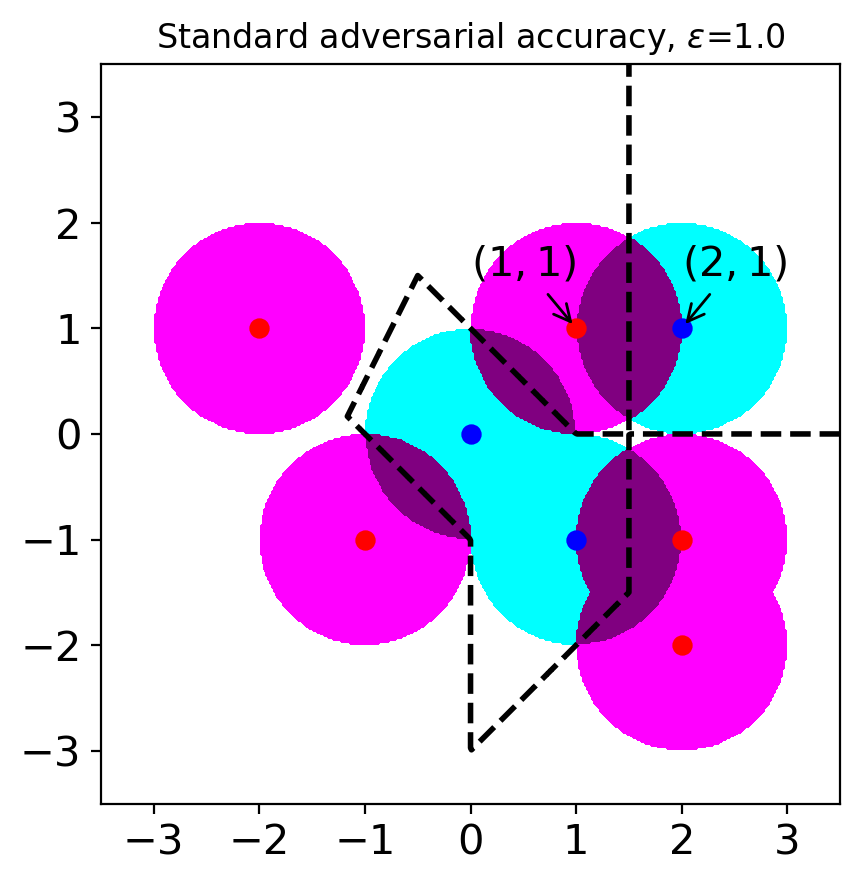} 
        \caption{{}}\label{fig:toy_ex_SAA_c}
    \end{subfigure}
    \caption{(a): Plot of the two-dimensional toy example. 
    Data points are colored based on their classes (\textcolor{red}{class \(A\): red} and \textcolor{blue}{class \(B\): blue}).
    (b): Visualization of the adversary regions %
    for SAA when \(\epsilon=0.5\). The regions are colored differently depending on their classes (\textcolor{red}{class \(A\):} \textcolor{magenta}{magenta} and \textcolor{blue}{class \(B\):} \textcolor{cyan}{cyan}). The decision boundary of a single nearest neighbor classifier is shown as a dashed black curve. (c): Visualization of the adversary regions %
    for SAA when \(\epsilon=1.0\). The overlapping adversary regions from the samples with different classes are colored in \textcolor{Plum}{purple}.}\label{fig:toy_ex_SAA}
\end{figure}

Let us consider an example visualized in \cref{fig:toy_ex_SAA_a}. %
The input space is \(\mathbb{R}^2\). There are only two classes \(A\) and \(B\), i.e., %
$\mathcal{Y}=\left\{ A, B\right\} $.
We use the \(l_2\) norm as a distance metric in this example. 

Let us consider a situation when \(\epsilon=1.0\) (see \cref{fig:toy_ex_SAA_c}). %
In this case, clean samples can also be considered as adversarial
examples. For example, the point \((2,1)\) can
be considered as an adversarial example
originating from the point \((1,1)\). If
we choose a robust model based on SAA,
we might choose a model with excessive
invariance. For example, we might choose a
model that predicts points belong to \(\mathbb{B}((1,1), 1)\) (including the point \((2,1)\)) have class A. Or, we can choose a
model that predicts points belong to \(\mathbb{B}((2,1), 1)\) (including the point \((1,1)\)) have class B. In either case, the accuracy of the chosen model is smaller than \(1\). This situation explains the tradeoff between accuracy and standard adversarial accuracy when large \(\epsilon\) is used. It originates from the overlapping adversary regions from the samples with different classes.

To avoid the tradeoff between accuracy and adversarial accuracy, one can use small \(\epsilon\) values.  %
Actually, a previous study has argued that commonly used \(\epsilon\) values are small enough to avoid the tradeoff \citep{yang2020closer}. However, when small \(\epsilon\) values are used, we can only analyze local robustness, and we need to ignore robustness beyond the chosen \(\epsilon\). 
For instance, let us consider our example when \(\epsilon=0.5\) (see \cref{fig:toy_ex_SAA_b}). 
In this case, we ignore robustness on \(\mathbb{B}((-2,1),1.0)-\mathbb{B}((-2,1),0.5)\). %
Models with local but %
without global robustness enable attackers to use large \(\epsilon\) values to fool the models. %
\citet{ghiasi2019breaking} have experimentally shown that even models with certified local robustness can be attacked by attacks with large $\epsilon$ values. Note that their attack applies little semantic perturbations even though the perturbation norms measured by \(l_p\) norms are large. %

These limitations motivate us to find an alternative way to measure robustness. \textbf{The contributions of this paper are as follows.}
\begin{itemize}
    \item We propose Voronoi-epsilon adversarial accuracy (VAA) that avoids the tradeoff between accuracy and adversarial accuracy. This allows the adversary regions to scale to cover most of the input space without incurring a tradeoff.
    To our best knowledge, this is the first work to achieve this without an external classifier. %
    (In \cref{sec:R_Vor}, we introduce formulas for adversary regions that can be used to estimate VAA.)
    \item We explain the connection between SAA and VAA. %
    We define global Voronoi-epsilon robustness %
    as a limit of the Voronoi-epsilon adversarial accuracy. We show that a nearest neighbor (1-NN) classifier maximizes global Voronoi-epsilon robustness. %
\end{itemize}

\section{Voronoi-Epsilon Adversarial Accuracy (VAA)}

Our approach restricts the allowed region of the perturbations to avoid the tradeoff originating from the definition of standard adversarial accuracy. This is achieved without limiting the magnitude of \(\epsilon\) and without using an external model. %
We want to have the following property to avoid the tradeoff. 
\begin{flalign}\label{eq:no_overlap_x}
\forall x_{i}, x_{j}\in X, \,\, x_{i}\neq x_{j} \Longrightarrow R(x_{i})\cap R(x_{j})=\varnothing
\end{flalign}

When Property (\ref{eq:no_overlap_x}) %
holds for the adversary region, we no longer have the tradeoff 
as \(x_i \notin R (x_{j})\)
for \(x_{i}\neq x_{j}\). In other words, a clean sample cannot be an adversarial example originating from another clean sample. We propose a new adversary called a Voronoi-epsilon adversary that combines the Voronoi-adversary introduced by \citet{khoury2019adversarial} with an \(\epsilon\)-ball-based adversary. This adversary is constrained to %
an adversary region \(Vor(x)\cap \mathbb{B}(x,\epsilon)\) where \(Vor(x)\) is the (open) Voronoi cell around a data point \(x\in X\). \(Vor(x)\) %
consists of every point in \(\mathcal{X}\) that is closer than any \(x_{clean}\in X-\left\{x\right\}\). Then, Property (\ref{eq:no_overlap_x}) holds as \(Vor(x_i)\cap Vor(x_j)=\varnothing\)
for \(x_{i}\neq x_{j}\). 

Based on a Voronoi-epsilon adversary,
we define Voronoi-epsilon adversarial accuracy (VAA). 

\begin{mydef}[\textbf{Voronoi-epsilon adversarial accuracy}]\label{def:vor_max_adv_acc}

When a Voronoi-epsilon adversary is used for the adversary, %
we refer to the adversarial accuracy %
as Voronoi-epsilon adversarial accuracy (VAA) $a_{Vor} (\epsilon)$. For VAA, we denote \(R({x})\) as \(R_{Vor}({\epsilon;x})\).

\begin{equation*}
    a_{Vor} (\epsilon)=\mathbb{E}_{x\in X}%
    \left[ {  \mathds{1}\left( f (x^{*})=c_x \right) } \right] %
    \text{ where } x^{*}=\argmax\limits_{x' \in R_{Vor}({\epsilon;x})}{L(x',c_x)}
\end{equation*}
\end{mydef}

Note that VAA is only defined on a fixed set of data points \(X\). %
As we do not know the distribution \(\mathcal{D}\), in practice, the fact that VAA is not defined on the whole input space does not matter.

\begin{figure}
    \centering
    \begin{subfigure}[b]{0.25\linewidth} %
        \includegraphics[width=1\linewidth]{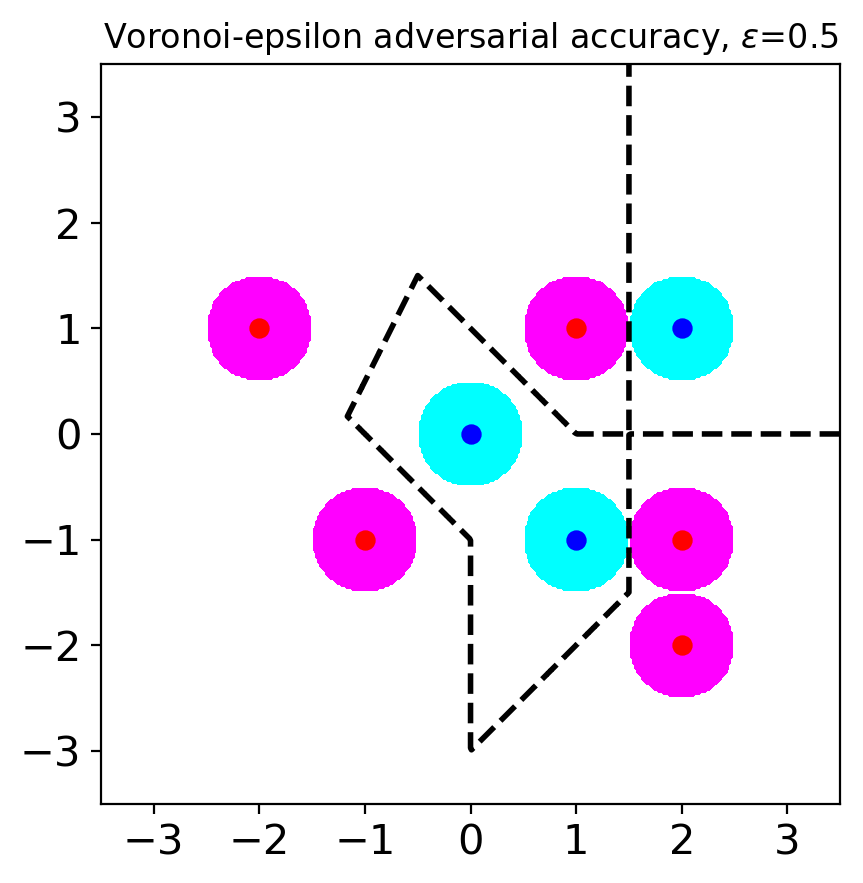}
        \caption{{}}\label{fig:toy_ex_VAA_a}
    \end{subfigure}
    \begin{subfigure}[b]{0.25\linewidth}%
        \includegraphics[width=1\linewidth]{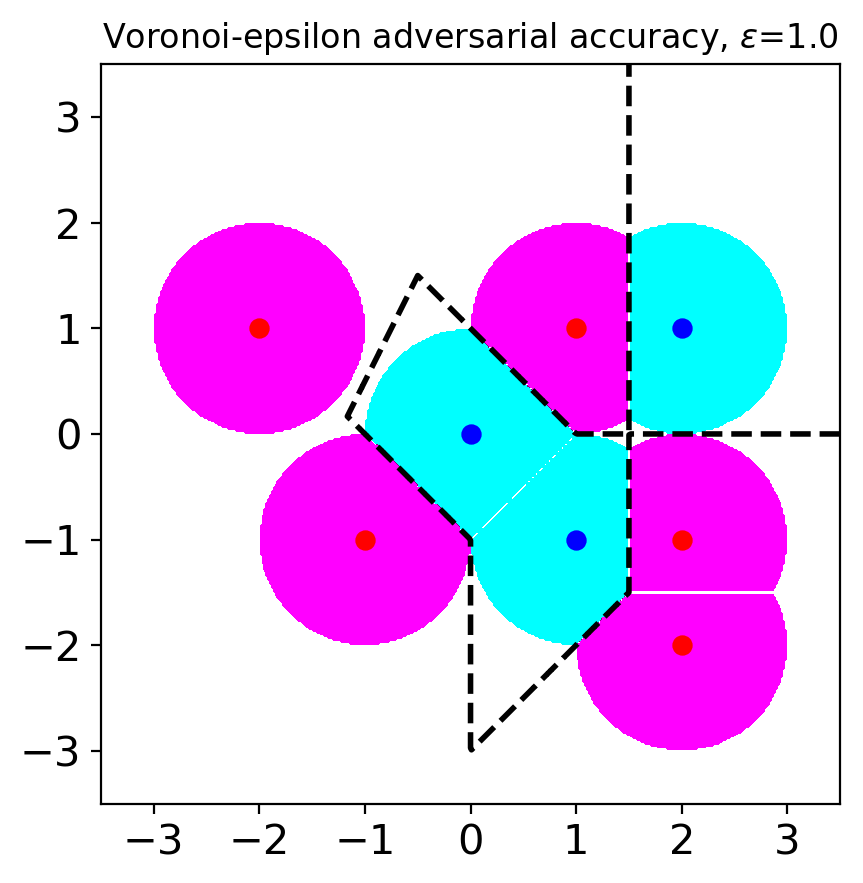} 
        \caption{{}}\label{fig:toy_ex_VAA_b}
    \end{subfigure}
    \begin{subfigure}[b]{0.25\linewidth}%
        \includegraphics[width=1\linewidth]{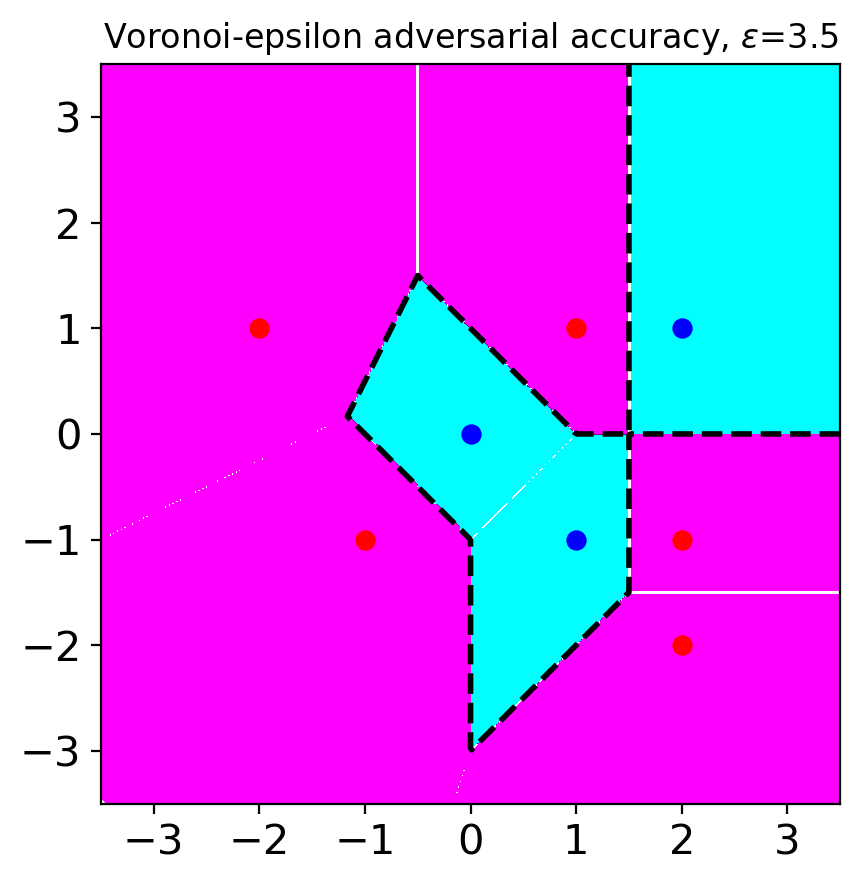} 
        \caption{{}}\label{fig:toy_ex_VAA_c}
    \end{subfigure}
    \caption{Visualization of the adversary regions  %
    for VAA with varying \(\epsilon\) values. The data points and regions are colored as in \cref{fig:toy_ex_SAA}. (a): When \(\epsilon=0.5\). %
    (b): When \(\epsilon=1.0\). (c): When \(\epsilon=3.5\).
    }\label{fig:toy_ex_VAA}
\end{figure}

\cref{fig:toy_ex_VAA} shows the adversary regions %
for VAA with varying \(\epsilon\) values. When \(\epsilon=0.5\), the regions are same with SAA except for the points \((1.5,1),(1.5,-1)\) and \((2,-1.5)\). %
Even when \(\epsilon\) is large (\(\epsilon>0.5\)), there is no overlapping adversary %
region, which was a source of the tradeoff in SAA. Therefore, when we choose a robust model based on VAA, we can get a model that is both accurate and robust. \cref{fig:toy_ex_VAA_c} shows the single nearest neighbor (1-NN) classifier would maximize VAA. The adversary regions %
cover most of the points in \(\mathbb{R}^{2}\) for large \(\epsilon\). %

\newtheorem{proposition}{Proposition}
\newtheorem{observation}{Observation}
\crefname{obs}{Observation}{Observations}
\Crefname{obs}{Observation}{Observations}

\begin{observation}\label{obs:VAA_SAA_eqivalence}
Let \(d_{min}\) be the nearest distance of the data point pairs, i.e., \(d_{min}=\min\limits_{x_i,x_j\in X, x_i\neq x_j} d(x_i,x_j)\). %
Then, the following equivalence holds.
\begin{flalign}\label{eq:VAA_SAA_equal}
a_{Vor} (\epsilon)=a_{std} (\epsilon) \text{ when } \epsilon<\frac{1}{2}d_{min}%
\end{flalign}
\end{observation}

Observation \ref{obs:VAA_SAA_eqivalence} shows that VAA is equivalent to SAA for sufficiently small \(\epsilon\) values. This indicates that VAA is an extension of SAA that avoids the tradeoff when \(\epsilon\) is large. %
The proof of the observation is in \cref{sec:VAA_calculation_proof}. We point out that equivalent findings were also mentioned in
\citet{yang2020robustness, yang2020closer, khoury2019adversarial}.

\newtheorem{theorem}{Theorem}

As explained in \cref{sec:toy_ex}, studying the local robustness of classifiers has a limitation. Attackers can attack models with only local robustness by using large \(\epsilon\) values. The absence of a tradeoff between accuracy and VAA enables us to increase \(\epsilon\) values and to study global robustness. We define a measure for global robustness using VAA. %

\begin{mydef}[\textbf{Global Voronoi-epsilon robustness}]\label{def:global_robustness} 
Global Voronoi-epsilon robustness \(a_{global}\) is defined as
\begin{equation*}
    a_{global}=\lim\limits_{ \epsilon \rightarrow \infty }{ a_{Vor} (\epsilon) }.
\end{equation*}
\end{mydef} 

Global Voronoi-epsilon robustness considers the robustness of classifiers for most points in \(\mathcal{X}\) 
(all points except for Voronoi boundary \(VB(X)\), which is the complement set of the unions of Voronoi cells.). We derive the following theorem from global Voronoi-epsilon robustness. %

\begin{theorem}\label{thm:1-NN_robust}
A single nearest neighbor (1-NN) classifier maximizes global Voronoi-epsilon robustness \(a_{global}\) on training data. %
1-NN classifier is a unique classifier that satisfies this except for Voronoi boundary \(VB(X)\).
\end{theorem}

Note that \cref{thm:1-NN_robust} only holds for exactly the same data under the exclusive class %
condition as mentioned in the problem settings
\ref{setting1}. It does not take into account generalization. %
The proof of the theorem is in \ref{sec:1-NN_robust_proof}.

\section{Discussion}
In this work, we address the tradeoff between accuracy and adversarial robustness by introducing the Voronoi-epsilon adversary. 
Another way to address this tradeoff is to use a Bayes optimal classifier \citep{suggala2019revisiting, kim2020sensible}. Since this is not available in practice, a reference model must be used as an approximation. 
In that case, the meaning of adversarial robustness is dependent on the choice of the reference model. %
VAA removes the need for a reference model by using the data point set \(X\) and the distance metric \(d\) to construct adversary. %
This is in contrast to \citet{khoury2019adversarial} who used Voronoi cell-based constraints (without \(\epsilon\)-balls) for an adversarial training purpose, but not for measuring adversarial robustness.

By avoiding the tradeoff with VAA, %
we can extend the study of local robustness to global robustness. 
Also, \cref{thm:1-NN_robust} implies that VAA is a measure of agreement with the 1-NN classifier. For sufficiently small \(\epsilon\) values, SAA is also a measure of agreement with the 1-NN classifier because SAA is equivalent to VAA as in Observation \ref{obs:VAA_SAA_eqivalence}. %
This implies that many defenses \citep{goodfellow2014explaining, madry2017towards, zhang2019theoretically, wong2018provable, cohen2019certified} %
with small \(\epsilon\) values unknowingly try to make locally the same predictions with a 1-NN classifier.

In our analysis, we do not consider generalization, and robust models are known to often generalize poorly \citep{raghunathan2020understanding}.  The close relationship between adversarially robust models and the 1-NN classifier revealed by \cref{thm:1-NN_robust} highlights a possible avenue to explore this phenomenon.

\subsubsection*{Acknowledgments}
We thank Dr. Nils Olav Handegard, Dr. Yi Liu, and Jungeum Kim for the helpful feedback. We also thank Dr. Wieland Brendel for the helpful discussions.

\bibliography{My_library} %
\bibliographystyle{iclr2021_conference}

\newpage

\appendix
\section{Appendix}
\subsection{List of Notation}\label{sec:notaion_list}
\begin{tabular}{p{1.25in}p{3.75in}}
$\displaystyle \epsilon$ & A perturbation budget.\\
$\displaystyle \dim$ & The dimension of the input space.\\
$\displaystyle \mathcal{X}$ & The nonempty input space. \(\mathcal{X} \subset \mathbb{R}^{\dim}\).\\
$\displaystyle \mathcal{Y}$ & The set of possible classes.\\
$\displaystyle c_{x}$ & A corresponding class of a clean data point \(x\in\mathcal{X}\).\\
$\displaystyle \mathcal{D}$ & The joint distribution. \(\mathcal{D}\subset \mathcal{X}\times\mathcal{Y}\).\\
$\displaystyle X$ & The set of data points. We assume it is a nonempty finite set.\\
$\displaystyle f$ & The classifier that we want to analyze. $f: \mathcal{X} \rightarrow \mathcal{Y}$.\\
$\displaystyle L(x,y)$ & A classification loss of the classifier \(f\) provided an input \(x\in \mathcal{X}\) and a label \(y \in \mathcal{Y}\).\\
$\displaystyle R({x})$ & An adversary region which is an allowed region of the perturbations for a data point \(x\). It can be depend on a perturbation budget \(\epsilon\).\\
$\displaystyle \mathds{1} \left( {} \right)$ & The indicator function. \(\mathds{1}\left(True\right)=1\) and \(\mathds{1}\left(False\right)=0\).\\
$\displaystyle a$ & Adversarial accuracy.\\
$\displaystyle d$ & The distance metric that is used for measuring adversarial robustness. It is not limited to \(l_p\) norms. It can be a learned metric or more complex distance.\\
$\displaystyle \mathbb{B}(x,\epsilon)$ & An \(\epsilon\)-ball around a sample \(x\). Mathematically, \(\mathbb{B}(x,\epsilon)=\left\{ {  x'\in\mathcal{X}}|{
d(x,x')
\le \epsilon } \right\} \).\\
$\displaystyle R_{std}({\epsilon;x})$ & The %
allowed regions of the perturbations 
for standard adversarial accuracy around a data point \(x\). \(R_{std}({\epsilon;x})=\mathbb{B}(x,\epsilon)\).\\
$\displaystyle a_{std} (\epsilon)$ & %
Standard adversarial accuracy using a perturbation budget \(\epsilon\). In other words, the adversarial accuracy when the adversary region is \(R_{std}({\epsilon;x})=\mathbb{B}(x,\epsilon)\).\\
$\displaystyle HS(x,x_{clean})$ & The (open) half-space closer to \(x\in X\) than \(x_{clean}\in X- \left\{ x \right\}\). Mathematically, \(HS(x,x_{clean})=\left\lbrace x'\in\mathcal{X} %
| d(x,x')<d(x_{clean},x')
\right\rbrace\).\\
$\displaystyle Vor(x)$ & The (open) Voronoi cell of a sample \(x\in X \). Mathematically, \(Vor(x)=\left\lbrace x'\in \mathcal{X}%
| d(x,x')<d(x_{clean},x')
, \forall x_{clean}\in X- \left\lbrace x  \right\rbrace \right\rbrace=\bigcap\limits_{x_{clean}\in X- \left\{ x \right\} }{HS(x,x_{clean})}\).\\
$\displaystyle R_{Vor}({\epsilon;x})$ & The allowed regions of the perturbations for Voronoi-epsilon adversarial accuracy around a data point \(x\). \(R_{Vor}({\epsilon;x})=Vor(x)\cap \mathbb{B}(x,\epsilon)\).\\
$\displaystyle a_{Vor} (\epsilon)$ & The Voronoi-epsilon adversarial accuracy using perturbation budget \(\epsilon\). In other words, the adversarial accuracy when the adversary region is \(R_{Vor}({\epsilon;x})=Vor(x)\cap \mathbb{B}(x,\epsilon)\). \\
$\displaystyle S^{\mathsf{c}}$ & The complement set of a set \(S\). For \(S\subset \mathcal{X}\), \(S^{\mathsf{c}}=\mathcal{X}-S\). \\
$\displaystyle VB(X)$ & Voronoi boundary based on \(X\). It is the complement set of the unions of Voronoi cells. \(VB(X)=\left(\bigcup\limits_{x\in X}{Vor(x)}\right)^{\mathsf{c}}=\bigcap\limits_{x\in X}{ Vor(x)^{\mathsf{c}} }\).\\
$\displaystyle a_{global}$ & Global Voronoi-epsilon robustness.\\
$\displaystyle N$ & The number of data points.\\
$\displaystyle R_{Vor;LB}({\epsilon;x})$ & The allowed regions of the perturbations for the lower bound of Voronoi-epsilon adversarial accuracy around a data point \(x\).  When \(\epsilon<\frac{1}{2}d(x,x_{m+2})\), \(R_{Vor;LB}({\epsilon;x})=R_{Vor}({\epsilon;x})\). When \(\epsilon\ge\frac{1}{2}d(x,x_{m+2})\), \(R_{Vor;LB}({\epsilon;x})=\mathbb{B}(x,\epsilon)\cap\left(\bigcap\limits_{i=2}^{m+1} {HS(x,x_i)}\right)\).%
\end{tabular}
\newpage
\begin{tabular}{p{1.25in}p{3.75in}}
$\displaystyle R_{Vor;UB}({\epsilon;x})$ & The allowed regions of the perturbations for the upper bound of Voronoi-epsilon adversarial accuracy around a sample \(x\). When \(\epsilon<\frac{1}{2}d(x,x_{m+2})\), \(R_{Vor;UB}({\epsilon;x})=R_{Vor}({\epsilon;x})\). %
When \(\epsilon\ge\frac{1}{2}d(x,x_{m+2})\), \(R_{Vor;UB}({\epsilon;x})=\mathbb{B}(x,\frac{1}{2}d(x,x_{m+2})
-\tau)\cap\left(\bigcap\limits_{i=2}^{m+1} {HS(x,x_i)}\right) \). \\
$\displaystyle a_{Vor;\, LB} (\epsilon)$ & The lower bound of Voronoi-epsilon adversarial accuracy using perturbation budget \(\epsilon\). It is defined as the adversarial accuracy when the adversary region %
for a data point \(x\) is \(R_{Vor;LB}({\epsilon;x})\).\\
$\displaystyle a_{Vor;\, UB} (\epsilon)$ & The upper bound of Voronoi-epsilon adversarial accuracy using perturbation budget \(\epsilon\). It is defined as the adversarial accuracy when the adversary region %
for a data point \(x\) is \(R_{Vor;UB}({\epsilon;x})\). %
\end{tabular}

\subsection{List of Abbreviation}\label{sec:abbreviation_list}
\begin{tabular}{p{1.25in}p{3.75in}}
$\displaystyle $AA & Adversarial accuracy.\\
$\displaystyle $SAA & Standard adversarial accuracy.\\
$\displaystyle $VAA & Voronoi-epsilon adversarial accuracy.\\
$\displaystyle $1-NN & Single nearest neighbor.\\
$\displaystyle $LB & Lower bound.\\
$\displaystyle $UB & Upper bound.
\end{tabular}

\subsection{Adversary Region
\texorpdfstring{$R_{Vor}({\epsilon;x})$}{R_Vor(epsilon;x )}}\label{sec:R_Vor} %
Voronoi-epsilon adversarial accuracy (VAA) uses \(R_{Vor}({\epsilon;x})=Vor(x)\cap \mathbb{B}(x,\epsilon)\). 
We introduce upper and lower bounds of \(R_{Vor}({\epsilon;x})\) using \(m+1\) nearest neighbors of a data point \(x\). These bounds enable to calculate approximate  upper and lower bounds of VAA.

\newtheorem{lemma}{Lemma}

\begin{lemma}\label{lem:allowed_region}
When \(N\) is the number of data points, let \(x_2,\cdots,x_{N}\in X-\left\lbrace x  \right\rbrace\) be the sorted neighbors of a data point \(x\in X\). Mathematically, 
\(d(x,x_2)\le d(x,x_3)\le \cdots \le d(x,x_{N})\).
Then, the following relations hold for a fixed number \(m\le N-2\).
\begin{fleqn} %
\begin{equation} \tag{3}\label{eq:allowed_region_1}
R_{Vor}({\epsilon;x})=\mathbb{B}(x,\epsilon) \text{ when } \epsilon<\frac{1}{2}d(x,x_2) 
\end{equation} 
\begin{equation} \tag{4}\label{eq:allowed_region_2}
\begin{split} 
R_{Vor}({\epsilon;x})=\mathbb{B}(x,\epsilon)\cap\left(\bigcap\limits_{i=2}^{j} {HS(x,x_i)} \right) 
\text{ when } \frac{1}{2}
d(x,x_{j})\le \epsilon<\frac{1}{2}d(x,x_{j+1}) \\
(j=2,\cdots,m+1) 
\end{split}
\end{equation} 
\begin{equation} \tag{5}\label{eq:allowed_region_3}
\begin{split}
\mathbb{B}(x,\frac{1}{2}d(x,x_{m+2})
-\tau)\cap\left(\bigcap\limits_{i=2}^{m+1} {HS(x,x_i)}\right) 
\subset R_{Vor}({\epsilon;x})\subset \mathbb{B}(x,\epsilon)\cap\left(\bigcap\limits_{i=2}^{m+1} {HS(x,x_i)}\right) \\
\text{ when } \epsilon\ge\frac{1}{2}d(x,x_{m+2})
\text{ and } \tau>0 
\end{split}
\end{equation} 
\end{fleqn} %
\end{lemma}

\begin{figure}[ht]
    \centering
    \begin{subfigure}[b]{0.225\linewidth}%
        \includegraphics[width=1\linewidth]{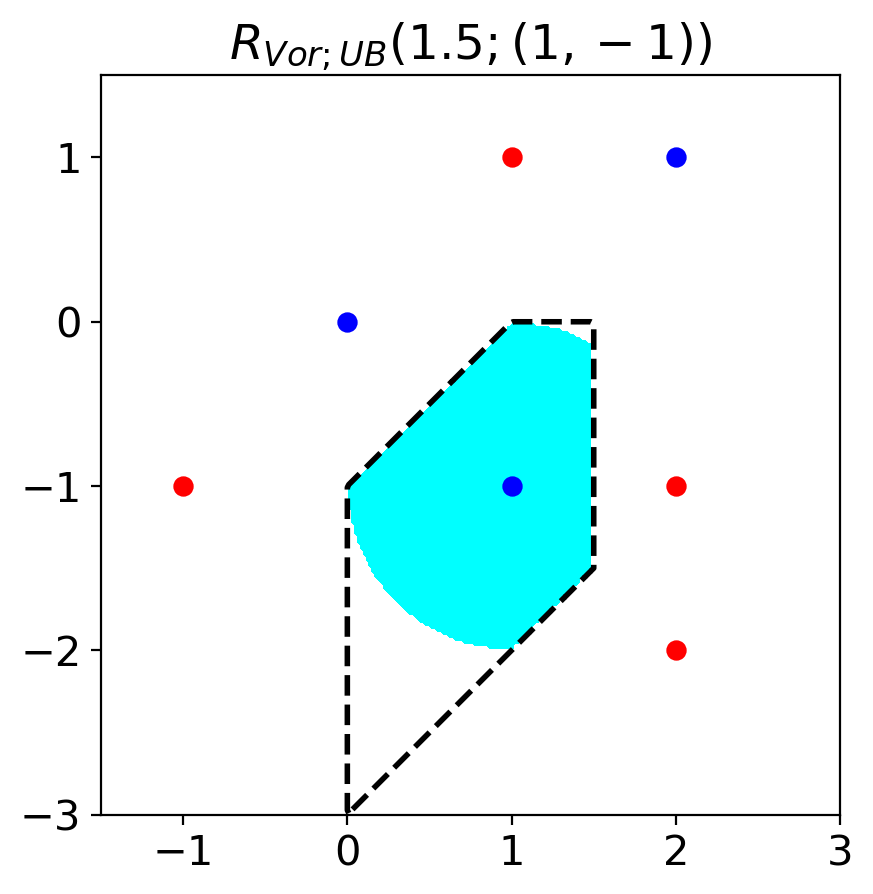}
        \caption{{}}\label{fig:toy_ex_regions_UB}
    \end{subfigure}
    \begin{subfigure}[b]{0.225\linewidth}%
        \includegraphics[width=1\linewidth]{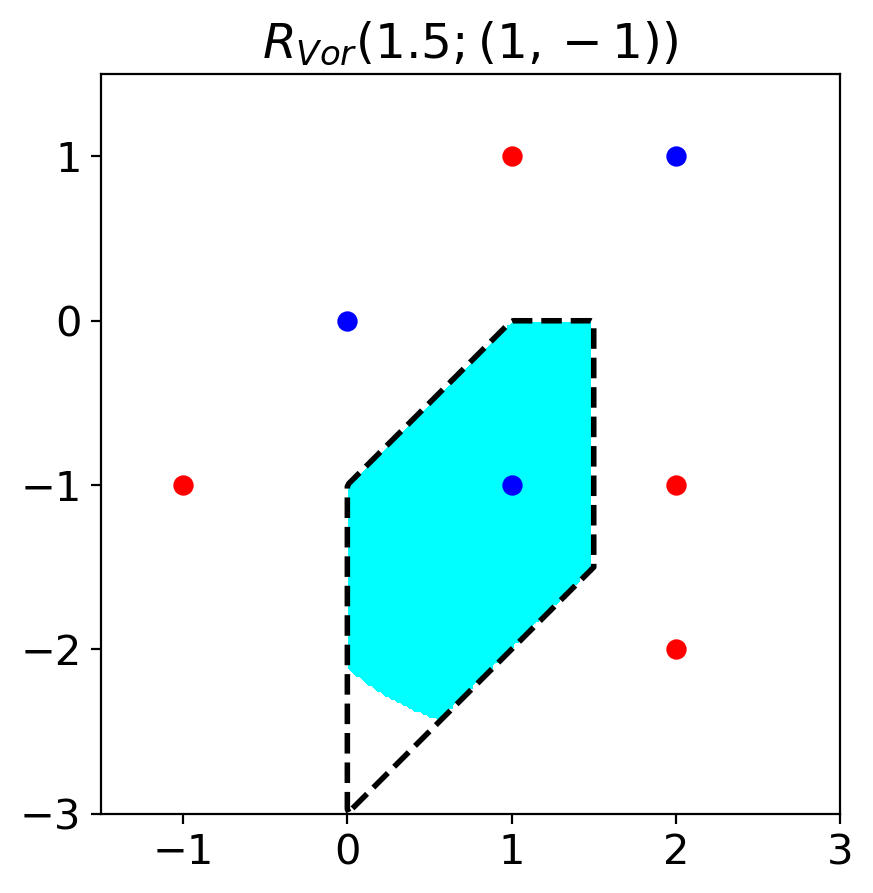}
        \caption{{}}\label{fig:toy_ex_regions_gen}%
    \end{subfigure}
    \begin{subfigure}[b]{0.225\linewidth}%
        \includegraphics[width=1\linewidth]{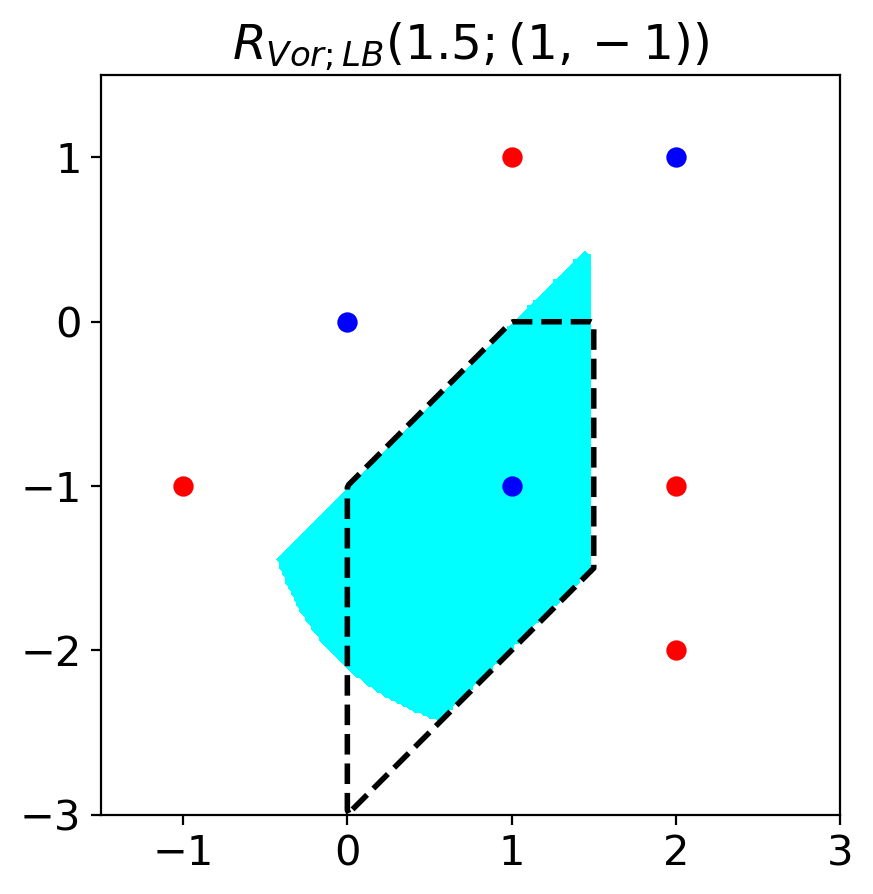}
        \caption{{}}\label{fig:toy_ex_regions_LB}
    \end{subfigure}
    \begin{subfigure}[b]{0.225\linewidth} %
        \includegraphics[width=1\linewidth]{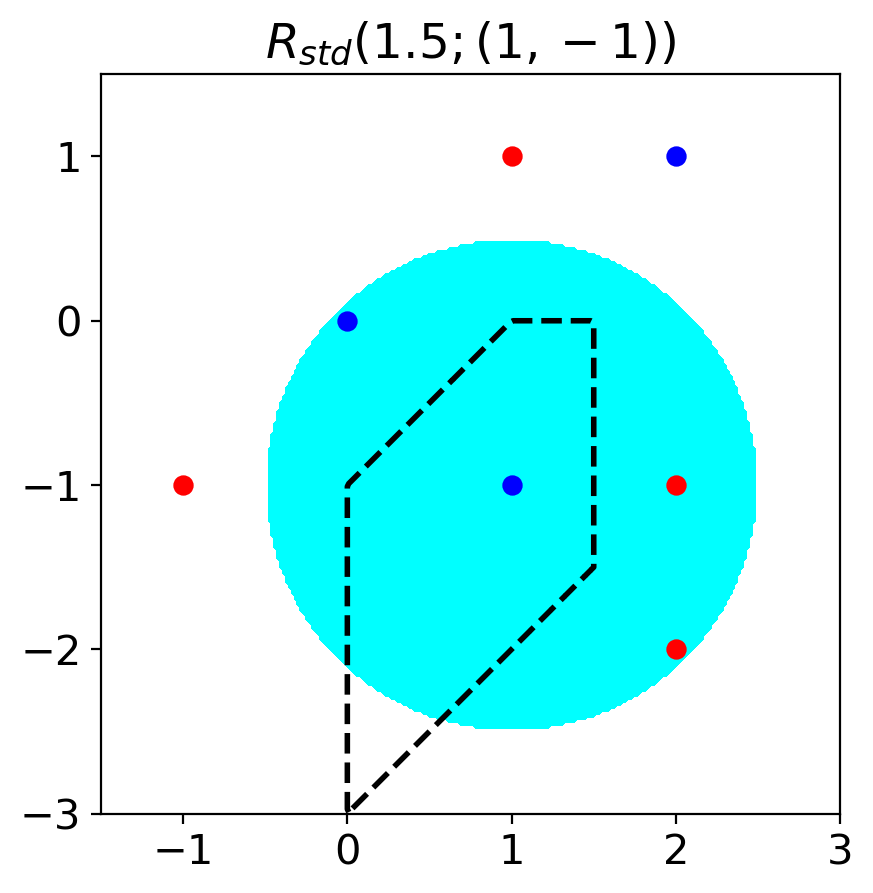}
        \caption{{}}\label{fig:toy_ex_regions_std}
    \end{subfigure}
    \caption{Visualization of the adversary region %
    for the point \((1,-1)\) when \(m=3\) and %
    \(\epsilon=1.5\) on our 
    example \ref{sec:toy_ex}.  %
    (a): \(R_{Vor;UB} (1.5;(1,-1))\). (b): \(R_{Vor} (1.5;(1,-1))\). (c): \(R_{Vor;LB} (1.5;(1,-1))\). (d): \(R_{std} (1.5;(1,-1))\).}\label{fig:toy_ex_regions}
\end{figure}

When \(\epsilon<\frac{1}{2}d(x,x_{m+2})\), we can calculate VAA using relations (\ref{eq:allowed_region_1}) and (\ref{eq:allowed_region_2}). The relation (\ref{eq:allowed_region_3}) of \cref{lem:allowed_region} enables to calculate the lower and upper bound of VAA when \(\epsilon\ge\frac{1}{2}d(x,x_{m+2})
\). When \(\epsilon\ge\frac{1}{2}d(x,x_{m+2})\), we denote the leftmost set in the relation (\ref{eq:allowed_region_3}) as \(R_{Vor;UB}({\epsilon;x})\) and the rightmost set as \(R_{Vor;LB}({\epsilon;x})\). (When \(\epsilon<\frac{1}{2}d(x,x_{m+2})\), we set \(R_{Vor;LB}({\epsilon;x})=R_{Vor;UB}({\epsilon;x})=R_{Vor}({\epsilon;x})\).) Figure \ref{fig:toy_ex_regions} visualizes the relationship \(R_{Vor;UB}({\epsilon;x})\subset R_{Vor}({\epsilon;x})\subset R_{Vor;LB}({\epsilon;x})\subset R_{std}({\epsilon;x})\). The proof of the lemma is in \cref{sec:allowed_region_proof}.

\begin{proposition}\label{prop:VAA_calculation}
$a_{Vor;\, LB} (\epsilon)$ is defined as the adversarial accuracy when the allowed regions of perturbation is \(R_{Vor;LB}({\epsilon;x})\). $a_{Vor;\, UB} (\epsilon)$ is defined as the adversarial accuracy when the allowed regions of perturbation is \(R_{Vor;UB}({\epsilon;x})\). Then, the following relation holds. 
\begin{flalign}
a_{std} (\epsilon)
\le
a_{Vor;\, LB} (\epsilon)
\le a_{Vor} (\epsilon)\le a_{Vor;\, UB} (\epsilon) %
\tag{6}\label{eq:VAA_calculation_2}
\end{flalign}
\end{proposition}

The proof of \cref{prop:VAA_calculation} is in \cref{sec:VAA_calculation_proof}.

\subsection{Proof of \cref{lem:allowed_region}}\label{sec:allowed_region_proof}
\begin{proof} 
\textbf{Relation (\ref{eq:allowed_region_1})}\\
First, we consider when \(\epsilon<\frac{1}{2}d(x,x_{2})
\).\\
Let \(x'\in \mathbb{B}(x,\epsilon)\). Then, \(d(x,x')
\le\epsilon\).\\
\(\frac{1}{2}d(x,x_2)
\le \frac{1}{2} d(x,x_{clean})
, \forall x_{clean}\in X-\left\{ x \right\} \).\\
Due to the triangle inequality, \(\frac{1}{2} 
d(x,x_{clean})
\le \frac{1}{2} d(x,x')
+\frac{1}{2} d(x',x_{clean})
\).\\
When we combine the above inequalities, \(d(x,x')
\le \epsilon<\frac{1}{2}d(x,x_{2})
\le \frac{1}{2}d(x,x_{clean})
\le \frac{1}{2} d(x,x')
+\frac{1}{2} d(x',x_{clean})
, \forall x_{clean}\in X-\left\{ x \right\}\).\\
Then, \(\frac{1}{2} d(x,x')
<\frac{1}{2} d(x',x_{clean})=\frac{1}{2} d(x_{clean},x')
, \forall x_{clean}\in X-\left\{ x \right\}\). Thus, \(x'\in Vor(x)\). \\
Hence, \(\mathbb{B}(x,\epsilon)\subset Vor(x)\) and \(R_{Vor}({\epsilon;x})=\mathbb{B}(x,\epsilon)\cap Vor(x)=\mathbb{B}(x,\epsilon)\).\\
\textbf{Relation (\ref{eq:allowed_region_2})}\\
Now, we consider when \(\frac{1}{2}d(x,x_{j})
\le \epsilon<\frac{1}{2}d(x,x_{j+1})
\,(j=2,\cdots,m+1)\).\\
\(R_{Vor}({\epsilon;x})=\mathbb{B}(x,\epsilon)\cap\left( \bigcap\limits_{i=2}^{N-1} {HS(x,x_i)}\right) 
\subset\mathbb{B}(x,\epsilon)
\cap\left( \bigcap\limits_{i=2}^{j} {HS(x,x_i)}\right) \)
is obvious as \(j\le N-1\).\\ %
We only need to proof \(\mathbb{B}(x,\epsilon)\cap\left( \bigcap\limits_{i=2}^{j} {HS(x,x_i)}\right) \subset R_{Vor}({\epsilon;x})\).\\
Let \(x'\in\mathbb{B}(x,\epsilon)\cap\left( \bigcap\limits_{i=2}^{j} {HS(x,x_i)}\right) \). Then, \(d(x,x')
\le\epsilon, d(x,x')
<d(x_2,x')
,\cdots,d(x,x')
<d(x_j,x')
\).\\
\( \frac{1}{2}d(x,x_{j+1})
\le \frac{1}{2}d(x,x_{k})
\) for \(k=j+1,\cdots,N-1\).\\
Due to the triangle inequality, \(\frac{1}{2} d(x,x_{k})
\le \frac{1}{2} d(x,x')
+\frac{1}{2} d(x',x_k)
\).\\
When we combine the above inequalities, \(d(x,x')
\le \epsilon<\frac{1}{2}d(x,x_{j+1})
\le \frac{1}{2}d(x,x_{k})
\le \frac{1}{2} d(x,x')
+\frac{1}{2} d(x',x_k)
\) for \(k=j+1,\cdots,N-1\)
.\\
Then, \(\frac{1}{2} d(x,x')
<\frac{1}{2} d(x',x_k)=\frac{1}{2} d(x_k,x')
\) for \(k=j+1,\cdots,N-1\). \\
We got \(d(x,x')
\le\epsilon, d(x,x')
<d(x_2,x')
,\cdots,d(x,x')
<d(x_{N-1},x')
\) and we proved \(\mathbb{B}(x,\epsilon)\cap\left( \bigcap\limits_{i=2}^{j} {HS(x,x_i)}\right)\subset R_{Vor}({\epsilon;x})\).\\
\textbf{Relation (\ref{eq:allowed_region_3})}\\
Finally, we consider when  \(\epsilon\ge\frac{1}{2}
d(x,x_{m+2})
\).\\
(i)
\(\mathbb{B}(x,\frac{1}{2}d(x,x_{m+2})
-\tau)\cap\left(\bigcap\limits_{i=2}^{m+1} {HS(x,x_i)}\right) 
\subset R_{Vor}({\epsilon;x}) \text{ for }\tau>0\):\\
Let \(x'\in\mathbb{B}(x,\frac{1}{2}d(x,x_{m+2})
-\tau)\cap\left(\bigcap\limits_{i=2}^{m+1} {HS(x,x_i)}\right) \). Then, \(d(x,x')
\le\frac{1}{2}d(x,x_{m+2})
-\tau<\frac{1}{2}d(x,x_{m+2})
\le \epsilon, d(x,x')
<d(x_{2},x')
,\cdots,d(x,x')
<d(x_{m+1},x')
\).\\
Through similar process used in the proof of \textbf{Relation (\ref{eq:allowed_region_1})} and \textbf{Relation (\ref{eq:allowed_region_2})},  we have \(d(x,x')
<\frac{1}{2}d(x,x_{m+2})
\le \frac{1}{2}d(x,x_{k})
\le \frac{1}{2} d(x,x')
+\frac{1}{2} d(x',x_k)
\) for \(k=m+2,\cdots,N-1\).\\
Then, \(\frac{1}{2} d(x,x')
<\frac{1}{2} d(x',x_k)=\frac{1}{2} d(x_k,x')
\) for \(k=m+2,\cdots,N-1\). \\
We got \(d(x,x')
<\epsilon, d(x,x')
<d(x_2,x')
,\cdots,d(x,x')
<d(x_{N-1},x')
\) and we proved (i).\\
(ii)
\(R_{Vor}({\epsilon;x})\subset \mathbb{B}(x,\epsilon)\cap\left(\bigcap\limits_{i=2}^{m+1} {HS(x,x_i)}\right) \):\\
It is obvious as \(R_{Vor}({\epsilon;x})=\mathbb{B}(x,\epsilon)\cap\left( \bigcap\limits_{i=2}^{N-1} {HS(x,x_i)}\right)\) and \(m+1\le N-1\). 
\end{proof}

\subsection{Proof of Observation \ref{obs:VAA_SAA_eqivalence} and \cref{prop:VAA_calculation}}\label{sec:VAA_calculation_proof}
\begin{proof} 
\textbf{Observation \ref{obs:VAA_SAA_eqivalence}}\\ %
\(d_{min}\le d(x,x_i),\) \(\forall x,x_i\in X, x\neq x_i\).\\ 
When \(\epsilon<\frac{1}{2}d_{min}\), \(\epsilon<\frac{1}{2}d_{min}\le \frac{1}{2}d(x,x_i),\) \(\forall x,x_i\in X, x\neq x_i\). %
Thus, \(R_{Vor}({\epsilon;x})=\mathbb{B}(x,\epsilon),\,\forall x \in X\)  due to the relation (\ref{eq:allowed_region_1}) in \cref{lem:allowed_region}. \\
Then, 
\(a_{Vor} (\epsilon)\) is same with \(a_{std} (\epsilon)\) as \(R_{Vor}({\epsilon;x})=\mathbb{B}(x,\epsilon)=R_{std}({\epsilon;x})\)\(,\forall x\in X\).\\
\textbf{\cref{prop:VAA_calculation}}\\
First, we consider a data point \(x\in X\) and let \(x_2,\cdots ,x_{N}\in X-\left\lbrace x  \right\rbrace\) be the sorted neighbors of \(x\).\\
Let \(x^{*1}=\argmax\limits_{x' \in R_{std}({\epsilon;x})}{L(x',c_x)}\), \(x^{*2}=\argmax\limits_{x' \in R_{Vor;LB}({\epsilon;x})}{L(x',c_x)}\), \(x^{*3}=\argmax\limits_{x' \in R_{Vor}({\epsilon;x})}{L(x',c_x)}\), and \(x^{*4}=\argmax\limits_{x' \in R_{Vor;UB}({\epsilon;x})}{L(x',c_x)}\).\\
(i) When \(\epsilon<\frac{1}{2}d(x,x_{m+2})\):\\
\(R_{Vor;UB}({\epsilon;x})= R_{Vor}({\epsilon;x})= R_{Vor;LB}({\epsilon;x})\) from the definition.\\
\( R_{Vor;LB}({\epsilon;x})=R_{Vor}({\epsilon;x})\subset \mathbb{B}(x,\epsilon)=R_{std}({\epsilon;x})\) from the relations (\ref{eq:allowed_region_1}) and (\ref{eq:allowed_region_2}).\\
Then, \(\mathds{1}\left( f (x^{*1})=c_x \right)\le \mathds{1}\left( f (x^{*2})=c_x \right)= \mathds{1}\left( f (x^{*3})=c_x \right)=\mathds{1}\left( f (x^{*4})=c_x \right)\) as \(R_{Vor;UB}({\epsilon;x})=R_{Vor}({\epsilon;x})=R_{Vor;LB}({\epsilon;x})\subset R_{std}({\epsilon;x})\).\\ %
(ii) When \(\epsilon\ge\frac{1}{2}d(x,x_{m+2})\):\\
\(R_{Vor;UB}({\epsilon;x})\subset R_{Vor}({\epsilon;x})\subset R_{Vor;LB}({\epsilon;x})\) from the relation (\ref{eq:allowed_region_3}).\\
\( R_{Vor;LB}({\epsilon;x})=\mathbb{B}(x,\epsilon)\cap\left(\bigcap\limits_{i=2}^{m+1} {HS(x,x_i)}\right)\subset \mathbb{B}(x,\epsilon)=R_{std}({\epsilon;x})\) from the definition.\\ %
Then, \(\mathds{1}\left( f (x^{*1})=c_x \right)\le \mathds{1}\left( f (x^{*2})=c_x \right)\le \mathds{1}\left( f (x^{*3})=c_x \right)\le\mathds{1}\left( f (x^{*4})=c_x \right)\) as \(R_{Vor;UB}({\epsilon;x})\subset R_{Vor}({\epsilon;x})\subset R_{Vor;LB}({\epsilon;x})\subset R_{std}({\epsilon;x})\).\\
From (i) and (ii), \(\mathbb{E}_{(x,c_x)\sim \mathcal{D}} \left[ {  \mathds{1}\left( f (x^{*1})=c_x \right) } \right]\le \mathbb{E}_{(x,c_x)\sim \mathcal{D}} \left[ {  \mathds{1}\left( f (x^{*2})=c_x \right) } \right]\le \mathbb{E}_{(x,c_x)\sim \mathcal{D}} \left[ {  \mathds{1}\left( f (x^{*3})=c_x \right) } \right]\le \mathbb{E}_{(x,c_x)\sim \mathcal{D}} \left[ {  \mathds{1}\left( f (x^{*4})=c_x \right) } \right]\).\\
We finished the proof of the relation (\ref{eq:VAA_calculation_2})
as \(a_{std} (\epsilon)=\mathbb{E}_{(x,c_x)\sim \mathcal{D}} \left[ {  \mathds{1}\left( f (x^{*1})=c_x \right) } \right] \), \(a_{Vor;\, LB} (\epsilon)=\mathbb{E}_{(x,c_x)\sim \mathcal{D}} \left[ {  \mathds{1}\left( f (x^{*2})=c_x \right) } \right] \), \(a_{Vor} (\epsilon)=\mathbb{E}_{(x,c_x)\sim \mathcal{D}} \left[ {  \mathds{1}\left( f (x^{*3})=c_x \right) } \right] \), and \(a_{Vor;\, UB} (\epsilon)=\mathbb{E}_{(x,c_x)\sim \mathcal{D}} \left[ {  \mathds{1}\left( f (x^{*4})=c_x \right) } \right] \).
\end{proof}

\subsection{Proof of Theorem \ref{thm:1-NN_robust} %
}\label{sec:1-NN_robust_proof}

To proof Theorem \ref{thm:1-NN_robust}, we introduce the following lemma.
\begin{lemma}\label{lem:surjectivity}
By changing \(\epsilon\) and \(x\in X\), \(x'\) that satisfies \(x'\in R_{Vor}(\epsilon;x)\) can fill up  \(\mathcal{X}\) except for \(VB(X)\). In other words, \(VB(X)^{\mathsf{c}}=\mathcal{X}-VB(X)\subset \bigcup\limits_{\epsilon\ge 0}{\left (\bigcup\limits_{x \in X}{R_{Vor}(\epsilon;x)}\right )}\).
\end{lemma}

\begin{proof}\textbf{Lemma \ref{lem:surjectivity}}\label{prf:pt1}\\
Let \(x'\in VB(X)^{\mathsf{c}}\).\\
\(VB(X)^{\mathsf{c}}=\mathcal{X}-VB(X)=\mathcal{X}-\left(\bigcup\limits_{x\in X}{Vor(x)}\right)^{\mathsf{c}}=\mathcal{X} \cap \left(\bigcup\limits_{x\in X}{Vor(x)}\right)=\bigcup\limits_{x\in X}{Vor(x)}\).\\
\(\exists x \in X \text{ such that } x'\in Vor(x)\).\\ %
Let \(\epsilon^{*}=d(x,x')\). Then, \(d(x,x')\le \epsilon^{*}\) and \(x'\in Vor(x)\).\\
\(x' \in \mathbb{B}(x,\epsilon^{*})\cap Vor(x)=R_{Vor}(\epsilon^{*};x)\subset \bigcup\limits_{\epsilon\ge 0}{\left (\bigcup\limits_{x \in X}{R_{Vor}(\epsilon;x)}\right )}\).\\
We proved \(VB(X)^{\mathsf{c}}\subset \bigcup\limits_{\epsilon\ge 0}{\left (\bigcup\limits_{x \in X}{R_{Vor}(\epsilon;x)}\right )}\).
\end{proof}

Now, we proof Theorem \ref{thm:1-NN_robust}.
\begin{proof} %
\textbf{Part 1}\label{prf:pt1_prf}\\
First, we prove that a 1-NN classifier maximizes global Voronoi-epsilon robustness. We denote the 1-NN classifier as $f_{1-NN}$ and calculate its global Voronoi-epsilon robustness.\\
For a data point \(x\in X\), let \(x'\in R_{Vor}(\epsilon;x)=\mathbb{B}(x,\epsilon)\cap Vor(x)\).\\
\(x'\in Vor(x) \Longleftrightarrow d(x,x')<d(x_{clean},x'),\forall x\in X-\left\{ x \right\}\).\\
As \(x'\in R_{Vor}(\epsilon;x) \subset Vor(x)\), \(x\) is unique nearest data point in \(X\) and thus \(f_{1-NN}(x')=c_x\).\\
When \(x^{*}=\argmax\limits_{x' \in R_{Vor}({\epsilon;x})}{L(x',c_x)}\), \(a_{Vor} (\epsilon)=\mathbb{E}_{(x,c_x)\sim \mathcal{D}}\left[ {  \mathds{1}\left( f_{1-NN} (x^{*})=c_x \right) } \right]=\mathbb{E}_{(x,c_x)\sim \mathcal{D}}\left[ 1 \right]=1\).\\
\(a_{global}=\lim\limits_{ \epsilon \rightarrow \infty }{ a_{Vor} (\epsilon) }=\lim\limits_{ \epsilon \rightarrow \infty }{ 1 }=1\). Thus, $f_{1-NN}$ takes the maximum global Voronoi-epsilon robustness \(1\).

\textbf{Part 2}\\
Now, we prove that if $f^{*}$ maximizes global Voronoi-epsilon robustness, %
then $f^{*}$ becomes the 1-NN classifier %
except for %
Voronoi boundary $VB(X)$.\\ %
Let $f^{*1}$ be a function that maximizes global Voronoi-epsilon robustness.\\
From the last part of the part \hyperref[prf:pt1_prf]{1}, when we calculate global Voronoi-epsilon robustness of $f^{*1}$, it should satisfy the equation \(a_{global}=1\).\\
For a data point \(x\in X\) and \(\epsilon_1<\epsilon_2\), \(R_{Vor}(\epsilon_1;x)=\mathbb{B}(x,\epsilon_1)\cap Vor(x)\subset \mathbb{B}(x,\epsilon_2)\cap Vor(x)=R_{Vor}(\epsilon_2;x)\).\\ 
Thus, for a data point \(x\in X\) and \(\epsilon_1<\epsilon_2\), \(\mathds{1}\left( f^{*1} (x^{*1})=c_x \right)\ge \mathds{1}\left( f^{*1} (x^{*2})=c_x \right)\) where \(x^{*1}=\argmax\limits_{x' \in R_{Vor}({\epsilon_1;x})}{L(x',c_x)}\) and \(x^{*2}=\argmax\limits_{x' \in R_{Vor}({\epsilon_2;x})}{L(x',c_x)}\).\\
\( a_{Vor} (\epsilon_1)=\mathbb{E}_{(x,c_x)\sim \mathcal{D}} \left[ {  \mathds{1}\left( f^{*1} (x^{*1})=c_x \right) } \right]\ge \mathbb{E}_{(x,c_x)\sim \mathcal{D}} \left[ {  \mathds{1}\left( f^{*1} (x^{*2})=c_x \right) } \right]=a_{Vor} (\epsilon_2)\) for \(\epsilon_1<\epsilon_2\). In other words, \(a_{Vor} (\epsilon)\) is a decreasing function.\\
\(a_{Vor} (\epsilon)=1,\,\forall \epsilon\ge 0\) (\(\because a_{Vor} (\epsilon^{*})<1\) for a \(\epsilon^{*}>0\), then it is a contradictory to \(a_{global}=1\) as  \(a_{Vor} (\epsilon)\) is a decreasing function.).\\
\(1=a_{Vor} (\epsilon)=\mathbb{E}_{(x,c_x)\sim \mathcal{D}} \left[ {  \mathds{1}\left( f^{*1} (x^{*})=c_x \right) } \right]\) where \(x^{*}=\argmax\limits_{x' \in R_{Vor}({\epsilon;x})}{L(x',c_x)}\).\\
As the calculation is based on the finite set \(X\), \(f^{*1} (x^{*})=c_x\) (\(\because {\mathds{1}\left( f^{*1} (x^{*})=c_x \right) }=1\)) where \(x^{*}=\argmax\limits_{x' \in R_{Vor}({\epsilon;x})}{L(x',c_x)}\).\\
As \(x^{*}\) are the worst case adversarially perturbed samples, i.e., samples that output mostly
different from \(c_x\), \(f^{*1} (x')=c_x=f_{1-NN} (x')\) where \(x'\in R_{Vor}({\epsilon;x})\).\\
By changing \(\epsilon\) and \(x\in X\), \(x'\) that satisfies \(x'\in R_{Vor}({\epsilon;x})\) can fill up \(\mathcal{X}\) except for
\(V B(X)\) (\(\because\) \cref{lem:surjectivity}). Hence, \(f^{*1}\) is equivalent to \(f_{1-NN}\) except
for Voronoi boundary \(V B(X)\).\\ %

\end{proof}

\newcolumntype{M}[1]{>{\centering\arraybackslash}m{#1}}

\end{document}